%% file: acl_latex.tex
\pdfoutput=1
\documentclass[11pt]{article}

\usepackage[final]{acl}
\usepackage{times}
\usepackage{latexsym}

\usepackage[T1]{fontenc}
\usepackage[utf8]{inputenc}

\usepackage{microtype}

\usepackage{inconsolata}

\usepackage{graphicx}

\usepackage{booktabs}
\usepackage{multirow}
\usepackage{caption}
\usepackage{float}

\usepackage[switch]{lineno}

\begin{document}

\title{Navigating Tomorrow: Reliably Assessing Large Language Models Performance on Future Event Prediction}

\author{Petraq Nako \\
  University of Innsbruck \\
  Innsbruck, Austria \\
  \texttt{petraq.nako@student.uibk.ac.at} 
  \\\And
  Adam Jatowt \\
  University of Innsbruck \\
  Innsbruck, Austria \\
  \texttt{adam.jatowt@uibk.ac.at}}

\maketitle

\begin{abstract}
\input{sections/abstract} 
\end{abstract}

\section{Introduction}
\label{Introduction}
\input{sections/introduction}

\section{Related Work}
\label{Background and Related Work}
\input{sections/background}

\section{Dataset}
\label{Dataset}
\input{sections/dataset}

\section{LLM Forecasting Analysis}
\label{Methodology}
\input{sections/methodology}

\section{Findings and Discussion}
\label{findings}
\input{sections/discussion}

\section{Conclusion}
\label{conclusion}
\input{sections/conclusion}

\section*{Limitations}

Throughout this study, several limitations and challenges shaped our approach and findings.
The initial challenge was the limited background research specifically focused on future prediction tasks, which required us to adapt broader temporal reasoning literature, potentially introducing inconsistencies. 
During the dataset creation process, we faced significant obstacles due to restricted access to news archives, making it difficult to find plausible events for future prediction tasks.
Large Language Models' ethical guardrails led to guarded responses, often defaulting to "no" for safety. This conservative approach impacted predictive capabilities research, leading to skewed results. Finally, there is a risk of bias incorporated in news articles that could take diverse forms \cite{farber2020multidimensional}.
%Furthermore, our hypotheses were not conclusively proven. The "before vs. after" analysis showed no significantly better performance in the before category, and the popular entities did not outperform unpopular ones, leaving us with inconsistent findings.

%In summary, our study faced numerous challenges, from dataset creation to understanding model mechanics. While these limitations impacted our findings, they
%also highlight areas for future research and development, emphasizing the need for balanced datasets, access to more powerful models, and deeper insights into model reasoning.

\section*{Ethics Statement}
Our research leverages the GPT-3.5 turbo model, and other LLMs.
%licensed under both the OpenAI License and the Apache-2.0 license, and the LLaMA model, distributed under Meta's LLAMA 2 Community License Agreement. 
We strictly adhere to the conditions set forth by these licenses. The datasets we use are sourced from repositories that permit academic use. To encourage ease of use and modification by the research community, we are releasing the artifacts developed during our study under the MIT license. Throughout the project, we have ensured that data handling, model training, and dissemination of results comply with all relevant ethical guidelines and legal requirements.

An important point to make is that using LLMs for forecasting needs to be accompanied by a careful consideration of the potential risks that may arise from acting on the generated forecasts. This applies not only to using LLMs but also to employing any forecasting tool.

Questions about future may have large range of diverse outcomes due to the large space of possible answers and their inherent uncertainty which can make them somehow more difficult to test, control and align with human values and expectations. 
While this may be used to achieve sub-optimal performance, we do not believe this is a suitable attack vector to achieve harmful behavior.

\bibliography{acl_latex}

%\newpage
%\appendix
%\onecolumn

\section*{Appendix}
\begin{figure*}[htb]
\includegraphics[width=2\columnwidth]{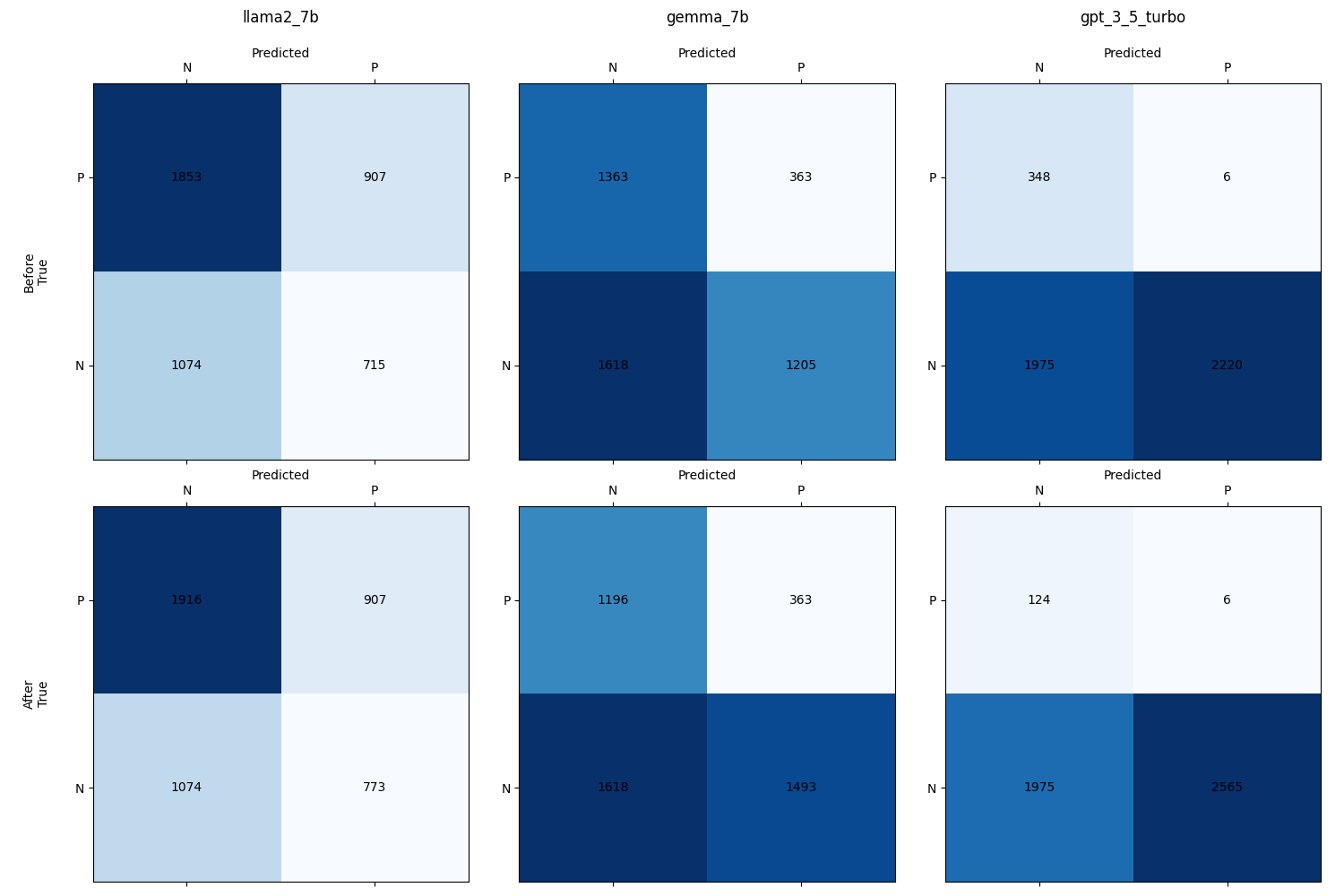} 
  \centering
  \caption{Confusion matrices of the \textbf{Before vs. After} categorization based on \texttt{Llama2 70b}, \texttt{Gemma 7b}, and \texttt{GPT 3.5 Turbo} models.}
  \label{fig:bef_aft_con_mat}
\end{figure*}

Figure \ref{fig:bef_aft_con_mat} displays confusion matrices of different models on the scenarios of before (upper row) and after (bottom row) cut-off dates of LLMs.

\begin{figure*}[]
\includegraphics[width=2\columnwidth]{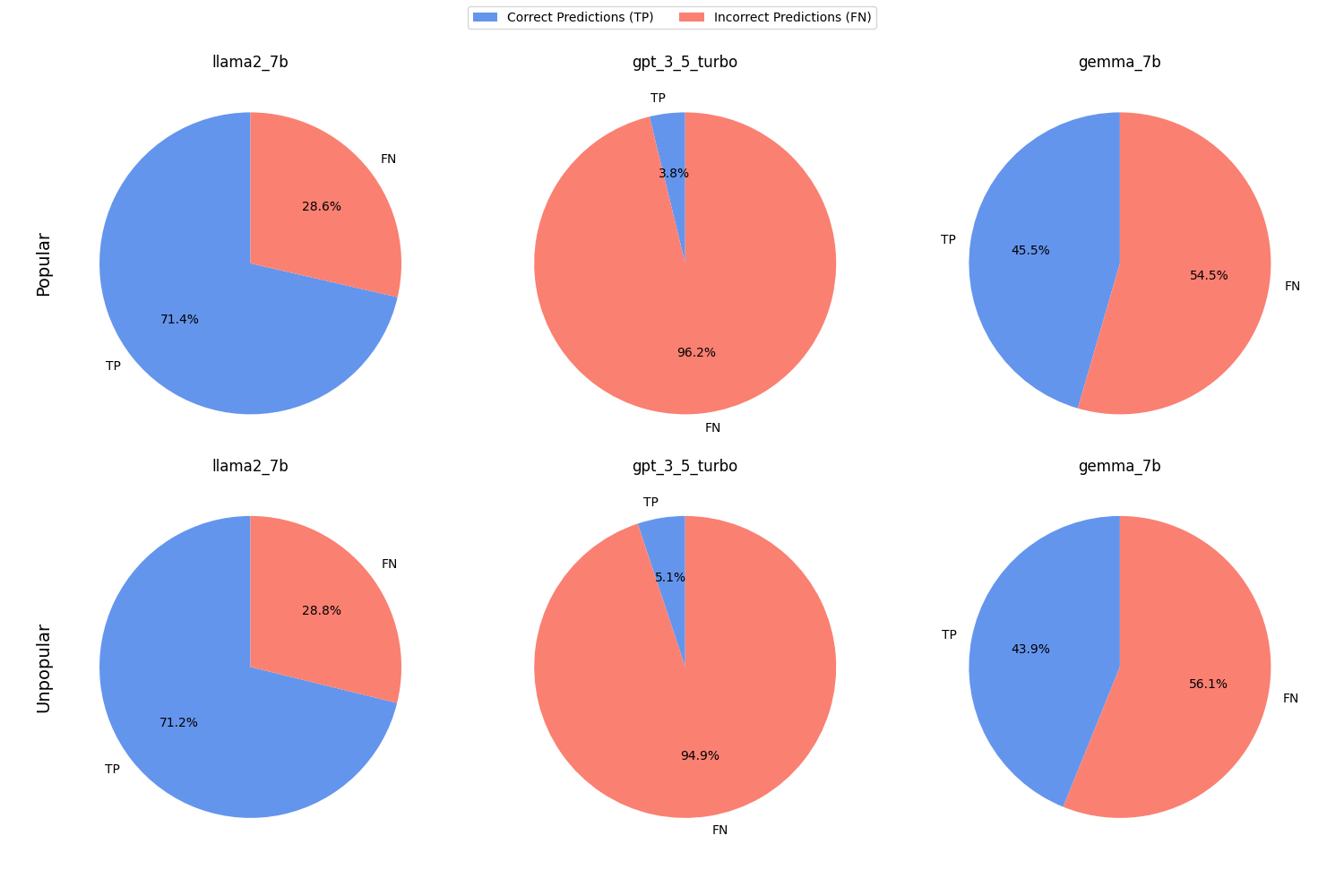} 
  \centering
  \caption{Performance comparison of the \textbf{Popular vs. Unpopular} categorization based on \texttt{Llama2 70b}, \texttt{GPT 3.5 Turbo} and \texttt{Gemma 7b} 
 models.}
  \label{fig:pop_pie}
\end{figure*}

Figure \ref{fig:pop_pie} provides performance comparison of models with respect to the numbers of correct and incorrect predictions for the cases of popular entities (upper row of plots) and unpopular entities (bottom row of plots).

%\label{sec:appendix_a}

\end{document}

%% file: sections/abstract.tex
Predicting future events is an important activity with applications across multiple fields and domains. For example, the capacity to foresee stock market trends, natural disasters, business developments, or political events can facilitate early preventive measures and uncover new opportunities. Multiple diverse computational methods for attempting future predictions, including predictive analysis, time series forecasting, and simulations have been proposed. This study evaluates the performance of several large language models (LLMs) in supporting future prediction tasks, an under-explored domain. We assess the models across three scenarios: Affirmative vs. Likelihood questioning, Reasoning, and Counterfactual analysis.
For this, we create a dataset\footnote{The dataset and all the code will be released after paper publication.} by finding and categorizing news articles based on entity type and its popularity. We gather news articles before and after the LLMs training cutoff date in order to thoroughly test and compare model performance. 
%Our findings indicate the Likelihood approach generally outperforms the Affirmative approach. Incorporating reasoning improves recall but increases false positives. Counterfactual analysis shows models' sensitivity to minor changes. 
Our research highlights LLMs' potential and limitations in predictive modeling, providing a foundation for future improvements.

%% file: sections/introduction.tex
Artificial Intelligence (AI) has made significant progress in recent years, particularly with the development of large language models (LLMs). Their use is however still underexplored across many complex tasks. One of them is supporting future prediction. Accurately predicting the future is crucial for anticipating and preparing for likely outcomes. This capability enables individuals to take essential actions, and allows authorities to develop necessary policies and make well-informed decisions. For instance, a company's business strategy and profitability heavily rely on their ability to forecast future trends effectively. Similarly, large organizations and governments continuously seek precise predictive tools and investors want to know most likely courses of actions before deciding to invest their money. In general, future forecasting is actually a quite common human activity, and LLMs deserve a closer investigation in this regard as a widely-used and disrupting technology.

Our research aims to understand the capability of LLMs in future forecasting and test diverse kinds of prompts to elucidate future-related content from parametric knowledge. 
%by evaluating the performance of various LLMs in different predictive scenarios. 
%The literature review includes works on temporal reasoning, future extraction, future prediction, and prompting strategies. 
For this, we first need to create a corresponding dataset. 
Prior investigations \cite{kanhabua2011ranking,jatowt2011Aextracting,jatowt2013multilingual} found that future-related information is relatively abundant in the Web, in particular, in news articles.
%approximately one-third of all sentences in news articles contain some type of references to the future. 
This lead to the recent creation of relevant datasets. The FORECASTQA dataset, as described in \citep{jin2020forecastqa}, involves collecting news articles from LexisNexis, filtering out non-English texts from 2015 to 2019, and converting them into <Question, Answer, Timestamp> triples to address binary and multiple-choice forecasting questions. However, FORECASTQA faces issues of ambiguity and the lack of context due to its crowdsourced nature. To overcome these challenges, \citet{zou2022forecasting} introduced the Autocast and IntervalQA datasets. Autocast includes True/False, Multiple-Choice, or Numerical forecasting questions, while IntervalQA comprises a large set of questions that only require numerical answers. 

The problems with the above-mentioned datasets are that they are not anchored in time and the analyses made by their authors do not respect time overlap. In consequence, it is unclear if the models tested on these datasets have not already seen the events to be forecasted. Hence, to assure trustworthy analysis, we create a dataset that clearly indicates the time of each question as well as the occurrence time of forecasted events, and we select LLMs whose training cut-off days are before the time when the events to be predicted occurred. We collect and categorize news articles by entity type and popularity, and then generate forecasting questions based on those events. This dataset helps evaluate underlying biases in the models and their forecast accuracy. We also split the news in our dataset into articles published before and after the LLMs' training dates to compare performance on probably "familiar" versus new information. Additionally, we generate negative instances to test the models' ability to distinguish real from fabricated events that did not occur. 

We then proceed to analyze LLM's capabilities in supporting forecasting of events. Our analysis involves three main scenarios: \textbf{Affirmative} vs. \textbf{Likelihood questioning}, \textbf{Reasoning}, and \textbf{Counterfactual analysis}. The Affirmative vs. Likelihood scenario compares the effectiveness of direct questions with likelihood-based questions. The Reasoning scenario examines whether incorporating logical reasoning into the prediction process can improve the models' performance. The Counterfactual analysis studies how sensitive the models are to slight changes in article details, testing LLMs' efficiency and adaptability.

Our findings reveal that
\emph{the Likelihood approach generally outperforms the Affirmative approach}, suggesting that probabilistic questioning provides a more nuanced understanding. \emph{Incorporating reasoning improves recall rates but increases false positives}, highlighting a trade-off between precision and recall. The \emph{Counterfactual analysis shows that models are sensitive to minor changes}, which significantly impact their performance.
In general, our analysis enhances the understanding of LLMs in predictive modeling and lays the groundwork for future improvements. 
%Future research should focus on improving model adaptability, developing better reasoning strategies, and creating more plausible datasets for prediction tasks.

%% file: sections/background.tex
Artificial intelligence has advanced significantly with the introduction of Large Language Models (LLMs), bringing us closer to machines that understand and communicate like humans. As these models develop, their potential goes beyond text generation to include temporal reasoning, future extraction, and future prediction.
%In particular, LLMs have been found to help with temporal reasoning. 
The field started with early foundational works such as McCarthy and Hayes' situation calculus \citep{McCarthy1981} that concentrated on representing and reasoning about change over time, and Allen's Interval Algebra \citep{Allen1983} which offered a framework for understanding relationships between time intervals, such as before, after, and during.
Then, later developments, including the TimeML framework \citep{Pustejovsky2003a} and datasets like TimeBank \citep{Pustejovsky2003b} integrated temporal reasoning into natural language processing (NLP). Evaluations using TIMEDIAL \citep{Qin2021TIMEDIAL} and TimeQA \citep{Chen2021TimeQA} revealed the limitations of LLMs in capturing the subtle details of everyday events. The TempEval challenges  \citet{Verhagen2007}, \citet{Verhagen2010} and \citet{UzZaman2013} further facilitated developments in temporal reasoning by providing benchmarks for evaluating temporal information extraction systems. The recent overview of temporal commonsense reasoning including also the use of LLM approaches is available in \cite{DBLP:journals/corr/abs-2308-00002}.

An important challenge for large language models (LLMs) is extracting future-related information from large amounts of textual data. \citet{regev2023futuretimelines} create an automated system for sifting through news articles to extract future-related content.  \citet{jatowt2011Aextracting} propose a text-clustering algorithm to extract collective future expectations from large text collections. \citet{jatowt2013multilingual} further the range to analyze future-related content in different languages, identifying cultural differences in future perception. \citet{kawai2010chronoseeker} proposed a search engine for both future and past events that expands a user query by some typical expressions related to event information such as year expressions, temporal modifiers and context terms, and filters out noisy events.
%They focused on content with a more future-oriented date. 
Other works include  \citet{Dias2014FutureRW}, who explored the task of future retrieval (retrieving documents containing future-related information pertinent to user queries), and  \citet{kanazawa2011}, who focused on improving information extraction for effective judgment of future outcomes. Additionally, \citet{radinsky2008} utilized web search patterns for predicting news occurrences, and \citet{nakajima2020} explored morphosemantic patterns for future predictions. Recent studies by \citet{zou2022forecasting} and \citet{kvamme2019timetoevent} demonstrated also the potential of neural networks in future prediction, while \citet{hu2017} introduced a context-aware model for generating short text predictions. Recently, \citet{regev2023futuretimelines} proposed to extend the forecasting task to constructing future timelines as a timeline summarization problem \cite{yu-etal-2021-multi}.

There has been few works on predicting future using LLMs \cite{li2024futurelanguagemodelingtemporal,gwak2024forecastingfutureinternationalevents,jin-etal-2021-forecastqa,Yuan2023BackFuture,zou2022forecasting}. 
\citet{jin-etal-2021-forecastqa} introduced FORECASTQA, which redefines event forecasting as a question-answering task. \citet{Yuan2023BackFuture} presented the ExpTime dataset in their "Back to the Future" study. According to the authors, the dataset improved LLMs' complex temporal reasoning and explainability capabilities. 

The existing studies involving LLMs are however still scarce while the available datasets are subject to contamination issue due to the temporal overlap, i.e., the lack of alignment of LLMs' cut-off dates and the dates of event occurrences. In this work, we attempt to overcome those challenges. We provide a temporally anchored dataset along with the set of LLMs with known cut-off dates that do not interfere with the dataset temporality. We also conduct an indepth analysis of different prompt formats to improve LLMs forecasting.

%Significant attention of the community has been also placed on model updating and performance on unseen entities or contents \cite{onoe2022entityclozedatelms,shi2024continuallearninglargelanguage}.

%% file: sections/dataset.tex
\begin{figure*}[t]
  \centering
  \includegraphics[height=0.35\textheight, keepaspectratio]{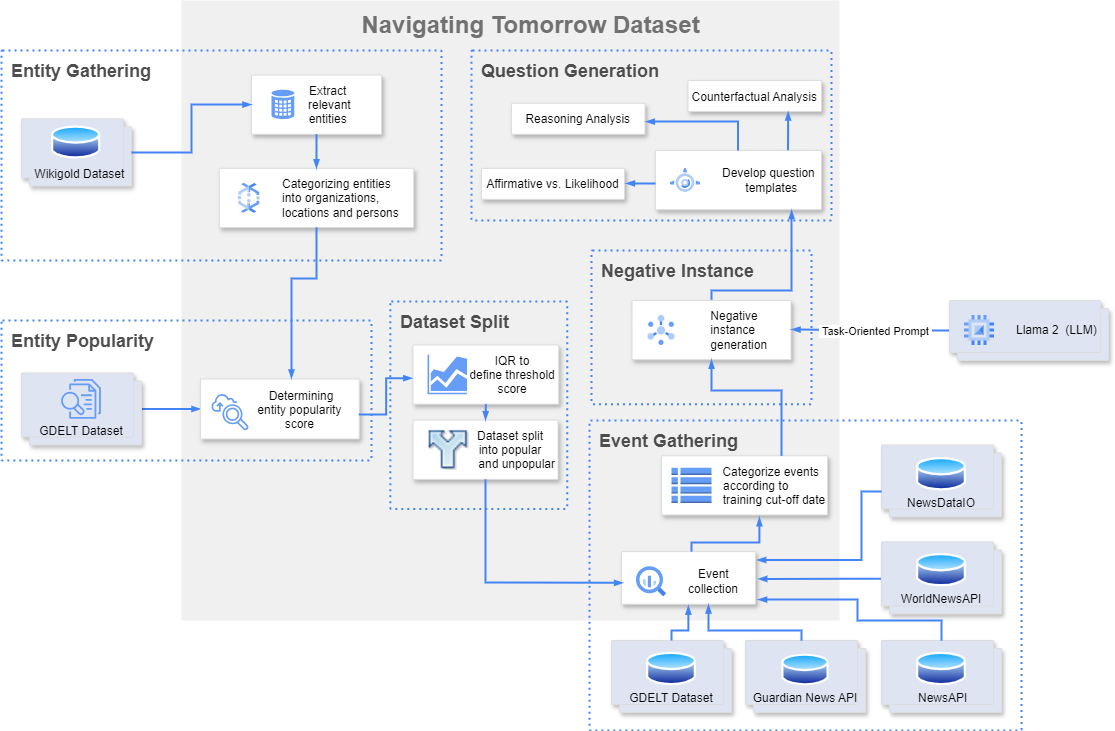} 
  \caption{The pipeline of our dataset creation.}
  \label{fig:dataset_pipeline}
\end{figure*}

We developed a specialized dataset to assess how well large language models performed in the forecasting tasks. Our approach involved several methodical steps, each created to ensure the dataset's relevance and comprehensiveness. Figure \ref{fig:dataset_pipeline} shows the pipeline of our dataset creation process, which we explain in detail in the following sections.

\subsection{Entity Gathering}

The initial phase involved utilizing the Wikigold dataset \citep{balasuriya-etal-2009-named}, a common resource for Named Entity Recognition (NER) tasks. Unlike conventional entity extraction processes, we did not extract entities directly from texts. Instead, we utilized the existing entity annotations provided in the Wikigold dataset, which includes a rich collection of pre-identified organizations, locations, and persons. These entities were then further adjusted and filtered to fit the specific requirements of our project, such as ensuring coverage within the context of predictive tasks. This approach eliminated the need for additional extraction models and focused on leveraging established resources for consistency. Wikigold dataset, which includes Wikipedia text, was chosen due to its representativeness of online content and the correspondence to Wikipedia page view analytics scores which we will use later. We collected available entities separating them into organizations, locations, and persons. The initial dataset included 898 organizations, 1,014 locations, and 934 persons. After data filtering based on assuring the availability of corresponding news articles, these numbers were further reduced to 652 organizations, 504 locations, and 478 persons.

\subsection{Determining Entity Popularity}

To evaluate how LLMs perform with entities of differing popularity, we measured the popularity of our entities using Wikipedia page view analytics \citep{wikimediaapi2023}. By aggregating monthly page views from January 2020 to December 2023, we assigned a popularity score to each entity. This step was taken to make it possible to analyze whether LLMs perform differently on "popular" vs. "unpopular" entities.
%The collected scores allowed us to analyze whether LLMs exhibited varying prediction accuracies for well-known versus lesser-known entities. 
For instance, prominent entities with consistently high page views may benefit from more extensive representation in training datasets, while lesser-known entities might expose gaps in LLM generalization capabilities.
In addition to calculating popularity scores, these analytics provided insights into broader trends, such as the relative visibility of different categories (e.g., organizations vs. persons). These patterns were critical in assessing potential biases in LLM predictions and their alignment with real-world prominence.

%\subsection{Dataset Split}

To create a balanced dataset, we categorized entities into popular and unpopular based on their popularity scores. Given the skewed distribution of popularity scores, we utilized the Interquartile Range (IQR) method to define the threshold score for this split. This approach ensured a reliable division by focusing on the middle 50\% of data, thus providing a reliable threshold for categorization.

\subsection{Event Collection}
\begin{figure}[t]
  \includegraphics[width=1.05\columnwidth]{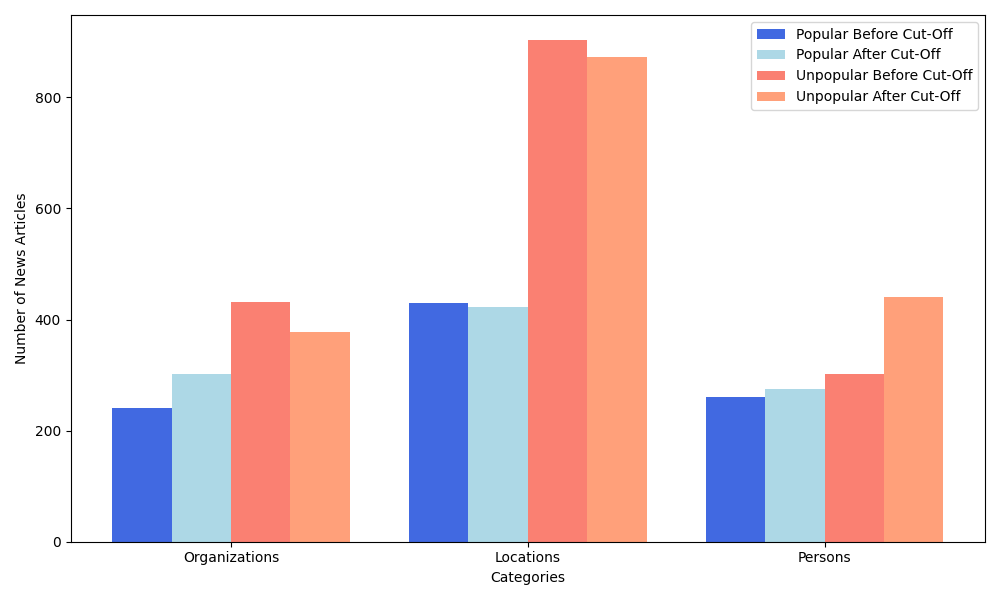} 
  \caption{News article distribution before and after cut-off training date.}
  \label{fig:event_distribution}
\end{figure}

One of the difficult parts of the dataset creation was gathering news articles about the events that were to be forecasted. Our search criteria were that the entity name should be mentioned in the news article's title. Articles that did not clearly reference the entities were excluded, ensuring that only relevant articles were included in the dataset. This filtering process helped eliminate noise and maintain the dataset's focus on meaningful relationships between entities and events. 
We collected articles published both before and after the LLMs' training cutoff dates to compare model performance on probably "familiar" versus novel events. Using free News API Providers like GDELT \citep{gdeltapi}, Guardian News API \citep{guardianapi}, NewsAPIAI \citep{newsapiai}, NewsAPIORG \citep{newsapiorg} and others, we collected articles about events related to the entities from our dataset. To manage the large volume of news articles collected during the dataset creation, we employed the Summarizer library 
%a pre-trained tool
which belongs to Transformers package and utilizes BERT \citep{Devlin2019BERT} for producing short extractive summaries. 
%This summarization tool was chosen for its ability to effectively generate extractive summaries, retaining the most relevant portions of the input text. 
The model was not fine-tuned specifically for the future prediction task and only relied on its pre-trained capabilities for contextual understanding and creating summaries. A random sample of the summaries was later manually reviewed to ensure they accurately captured the essence of events related to the selected entities.
%and focused on future events rather than past occurrences.
The final dataset consists of over 5,000 future event summaries about 657 entities (194 organizations, 288 locations, and 175 persons) evenly split between true and fake events. The creation of fake future events is explained in the next subsection.
Figure \ref{fig:event_distribution} provides a graphical overview of news article distribution by popularity and news article dates according to the LLMs' training cut-off date.

\subsection{Negative Instances}

To be able to test the models' ability to distinguish between real and fabricated events, we generated negative instances (fake news articles) using the \texttt{Llama2 7b-chat-hf} model \citep{2023llama2}. These articles were designed to mimic real news in style and content, creating a robust challenge for the models. The generation process involved task-oriented prompts such as: "Generate three fake news articles related to [ENTITY], each with a short summary (max three sentences) and a randomly chosen date in early 2023 (format: DD.MM.YYYY)."

Hyperparameters included a temperature of 0.5 to balance creativity and coherence. Iterative runs ensured diversity in the generated articles. To verify non-authenticity, generated content was manually reviewed and cross-checked against news databases to confirm that no real-world counterparts existed. This rigorous process ensured that fabricated articles were indistinguishable from real news in style while remaining fictional. 
%Additionally, a random sample of the generated fake news articles was later manually checked along with a random sample of true to verify their plausibility and to ensure they did not reference memorized real-world facts.

\subsection{Question Generation}

The final step involved formulating questions based on the event descriptions provided in the collected news articles. This was important for evaluating the models' predictive and reasoning capabilities using QA inputs. We first asked LLM (\texttt{GPT 3.5}) to generate a short (sentence-long) description (denoted as [EVENT]) of an event based on its summary. Based on those we then developed specific templates for different types of questioning:
\begin{enumerate}
    \item Affirmative vs. Likelihood Question Templates 
    \begin{itemize}
        \item \textbf{Affirmative Question Template}: "\textit{Will the following event [EVENT] happen on [DATE (news' published date written as Month, YYYY)]? Please only answer with yes or no.}"
        \item \textbf{Likelihood Question Template}: "\textit{Is it likely that the following event [EVENT] will occur in [DATE (news' published date written as Month, YYYY)]? Please only answer with yes or no.}"

    \end{itemize}
\item \textbf{Reasoning Question Template}
%\begin{itemize}
 %   \item 
    
    This template asks the models not only to predict the occurrence of an event but also to provide a rationale for their prediction. The prompt was structured as: "\textit{Is it likely that the following event [EVENT] will occur in [DATE (news' published date written as Month, YYYY)]? Please answer first only with yes or no. Then please explain shortly and concisely what made you decide on that answer.}"
%\end{itemize}
\item Counterfactual Question Template

For counterfactual analysis, we introduced two types of minor alterations to the event data:
\begin{itemize}
    \item \textbf{Temporal Adjustments}: The year associated with an event was shifted by adding or subtracting 2–3 years. For example, an event originally occurring in 2021 might be altered to 2019 or 2023. This adjustment tested whether the models would maintain consistent predictions despite slight changes in timing.
    \item  \textbf{Factual Alterations}: Certain details, such as financial figures or project scales, were modified in a controlled or random manner. For instance, "Company X reported a 20\% increase in profits" was altered to "Company X reported a 15\% decrease in profits." These changes were minimal enough to preserve the event's overall context but significant enough to test the models' sensitivity to factual variations. Importantly, such modifications were applied only to true events, while fabricated events (negative instances) were left unaltered to maintain their character.
\end{itemize}

\end{enumerate}

Prompts were primarily run once; however, in cases where models did not respond appropriately or failed to generate any answer, the questions were rerun until a valid response was obtained. Once a response was generated, we did not rerun the same question except for a small manual sample of the dataset. In these cases, the models consistently provided the same answers, indicating robustness for those specific instances. These templates were created to get different and detailed responses from the models, allowing us to comprehensively evaluate their ability to predict and reason. By using different types of questions, we aimed to see how well the models could make straightforward predictions, explain their reasoning, and respond to small changes in event details.

%% file: sections/methodology.tex
%Our methodology for evaluating the performance of large language models in forecasting tasks includes using several state-of-the-art models and applying a structured approach to question-answering and data analysis. This section outlines the rationale behind our chosen methods and the techniques used to analyze the models' performance. 

\subsection{Large Language Models}

We examine the question-answering capabilities of LLMs using a
variety of chosen models, each with distinct advantages based on their architectural
differences and the training cut-off dates. Understanding the specifications and
historical training context of these models is important for interpreting their performance and the relevance of their outputs to our dataset of news events, mostly
dating after June 2023.

\textbf{Llama2 7b and 70b} \citep{2023llama2}: Llama2 models, developed by Meta, are among the most advanced LLMs. With parameters varying from 7 billion to 70 billion, they are specifically tuned for dialogue-based tasks. Their training cut-off date is July 2023.

\textbf{GPT-3.5 Turbo} \citep{Radford2018GPT}: OpenAI's \citep{openai} GPT-3.5 Turbo is a generative pre-trained transformer model optimized for speed and coherence in conversation tasks. Utilizing diverse training data until September 2021, it is highly adaptable and capable of generating accurate responses across various topics and scenarios.

\textbf{Mistral 7b} \citep{jiang2023mistral}: Mistral 7b instruct is a high-performance model with 7 billion parameters, excelling in reasoning and code generation. It uses advanced techniques for faster interference and cost-effective sequence management. Its training cut-off date is September 2023.

\textbf{Mixtral 8x7b} \citep{jiang2024mixtral}: The Mixtral 8x7b model uses a Sparse Mixture of Experts (SMoE) architecture, with 8 feed-forward blocks per layer, allowing token interaction with different experts, optimizing performance and efficiency, and having the same training cut-off date as Mistral.

\textbf{Gemma 7b} \cite{gemmateam2024gemma}: Gemma 7b is part of Google’s effort to create accessible yet powerful
models that can be used in environments with limited computational
resources. Despite its smaller size compared to some of the other models used in this study, Gemma 7b does not compromise performance. Launched in February 2024, this model serves as a crucial benchmark within our research to assess how models trained with slightly earlier
data compare with those trained later in terms of understanding and predicting newer events

\subsection{LLMs Question-Answering}

To evaluate the predictive capabilities of the chosen LLMs, we designed three different question-answering approaches. Each approach was created to test different aspects of the models' performance, from straightforward predictions to more complex reasoning and adaptability to changes.

\textbf{Affirmative vs Likelihood Questioning}: This approach serves as the basic questioning strategy. It involves comparing direct affirmative questions with likelihood-based questions. The purpose is to determine which method generates more accurate predictions by assessing the models' ability to handle straightforward predictions versus probability-based evaluations. 

\textbf{Reasoning Analysis}: In this approach, models are asked not only to predict the occurrence of an event but also to provide an explanation for their prediction. This helps assess the models' ability to reason and articulate their thought process giving insight into the models' deeper understanding of the events and their ability to generate coherent explanations.

\textbf{Counterfactual Analysis}: This method tests the models' sensitivity to minor changes in event details by presenting them with slightly altered versions of the original events. The purpose is to evaluate how well the models can adapt to these variations and maintain accurate predictions. This approach is important for understanding the models' flexibility and robustness in dynamic scenarios.

\subsection{Data Analysis Techniques}

This section discusses the methodologies used to analyze data from large language model experiments, evaluating model accuracy and reliability, and understanding underlying patterns. Several statistical techniques and visualizations are used to provide a comprehensive view of both quantitative and qualitative aspects.

\textbf{Descriptive Statistics}: This method summarizes and organizes the dataset to give a clear overview of the models' responses. By categorizing responses by entity type (organizations, locations, persons) and popularity (popular, unpopular), we can analyze how the models perform across different groups. This helps identify patterns and trends in the data.

\textbf{Evaluation Metrics}: We utilize Precision, Recall, F1-Score, and Accuracy for
determining the accuracy and reliability of the models in contextual processing and
responding to our uniquely structured question templates.

%% file: sections/discussion.tex
\subsection{Findings}

In this section, we present the primary outcomes derived from utilizing large language models (LLMs) to tackle a set of designed question templates that test their predictive and reasoning capabilities. The analysis is structured around three principal scenarios: Affirmative vs. Likelihood Questions, Reasoning Analysis, and Counterfactual Analysis. 

\subsubsection{Affirmative vs Likelihood Analysis}

We evaluated model responses to both affirmative and likelihood questioning styles across various metrics to establish a baseline understanding of model performance. This analysis includes results both before (shown in Table \ref{tab:bef_aff_vs_lik}) and after (shown in Table \ref{tab:aft_aff_vs_lik}) the training cut-off date, examining how well the models could predict actual future events and their performance on events they may have "encountered" during training.

\input{tables/aff_vs_lik_before}

We use the \textbf{"before" } scenario to assess baseline performance and test the hypothesis that models would perform better on events they might have been exposed to during training. In the \textbf{"before cut-off"} events, the likelihood questioning approach generally resulted in higher Precision across most models. This suggests that the likelihood format leads to more accurate predictions. For example, the \texttt{Llama2 7b} model showed a Precision of 0.671 in the likelihood scenario, indicating its effectiveness in making accurate likelihood predictions. However, Recall rates were often higher for the affirmative approach, indicating that while the likelihood format is more precise, it is less inclusive in identifying positive events. The  \texttt{Llama2 70b} model, although it showed a lower Recall, demonstrated great Precision of 0.968, emphasizing its accuracy in predictions when it does classify an event as likely.

\input{tables/aff_vs_lik_after}

In the \textbf{"after cut-off"} scenario, intended to evaluate the true predictive capacity of the models, the preference for likelihood questioning was reinforced. For instance,  \texttt{Llama2 7b} maintained its lead with the highest Accuracy of 0.642, underscoring its efficiency in handling real predictive tasks. The model not only held its ground in Precision and Recall but also saw an increase in its F1-Score, indicating an even more balanced performance when facing true predictions of future events. Across both scenarios, the likelihood approach consistently resulted in higher Precision, highlighting its effectiveness in making correct predictions and reinforcing its suitability for predictive tasks where minimizing false positives is essential.

\subsubsection{Reasoning Analysis}

Following the analysis of model performance across the Affirmative vs. Likelihood scenarios, we turn our attention to the added dimension of reasoning within the likelihood questioning framework, focusing on the "after cut-off" events.
Recognizing the overall dominance of the Likelihood approach in initial assessments, all subsequent evaluations incorporate this format further to probe the models’ analytical depth and predictive accuracy. The comparative analysis, presented in Table \ref{tab:lik_and_reason}, examines how reasoning influences model responses. A notable aspect of this table is the presence of extreme values such as 0 and 1, in some metrics. These values arise from the models’ definitive responses to the reasoning prompts. The integration of reasoning capabilities into the likelihood approach results in a trade-off between Precision and Recall. For example,  \texttt{Llama2 7b} improves Recall but decreases Precision, leading to a marginal increase in false positives. Despite some instabilities in precision and recall, the addition of reasoning generally benefited the F1-Score, which balances these two metrics. For instance,  \texttt{Llama2 7b}’s F1-Score increased to 0.790, demonstrating a more effective balance between identifying true events and minimizing false positives.
\input{tables/reasoning}

\subsubsection{Counterfactual Analysis}

Lastly, we explore how slight factual modifications influence model predictions through counterfactual analysis (shown in Table \ref{tab:pos_vs_counter}). This analysis focuses on how minor changes, such as altering dates or minor facts in true events, impact model predictions. The results show a general decline in performance metrics for counterfactual instances compared to standard positive instances, suggesting that models struggle to adapt to slight changes. For instance, the \texttt{Llama2 7b} model showed a significant drop in performance when faced with counterfactual scenarios, highlighting potential difficulties in adapting to deviations from their training data.

\input{tables/counterfactual}

This raises the question of whether being highly sensitive to small changes is an advantage or a disadvantage. This question is important for understanding whether being highly affected by small changes is an advantage, making the models more robust and flexible, or a disadvantage, meaning the models cannot handle slight variations well. Identifying whether this sensitivity is a strength or a weakness requires further research, highlighting an important area for future research to ensure the models can be effectively tuned for real-world application.

\subsection{Interpretation of Results}

Following the findings from the last section, we dive deeper into the different aspects of the models’ performance.

\subsubsection{Before vs. After Categorization}
The comparison between the before and after scenarios reveals some interesting
trends. Figure \ref{fig:bef_aft_con_mat} in Appendix shows the confusion matrices based on   \texttt{Llama2 70b},  \texttt{Gemma 7b}, and  \texttt{GPT 3.5 turbo} models. The models’ performance metrics show a general decrease in the after categorization scenario, albeit a slight one. This indicates that the models’ ability to predict future events is not significantly worse than their performance on potential training data events. However, the expected better results in the before scenario are notably absent, which is a somewhat surprising discrepancy to our
initial hypothesis that the models would perform significantly better on events from their potential training data.

\subsubsection{Popularity Categorization}

Models demonstrated a slight preference for popular entities, but the difference in performance was not substantial. While popular entities showed marginally better Precision and Accuracy, this improvement was not consistent across all models. This suggests that the models handle both popular and unpopular entities with relatively equal proficiency, challenging the assumption that popularity significantly impacts performance.

Figure \ref{fig:pop_pie} in Appendix presents a visual overview of correct and incorrect predictions in the popularity categorization for the  \texttt{Llama2 70b}, \texttt{Gemma 7b}, and  \texttt{GPT 3.5 turbo} models.

\subsubsection{Entity Type Categorization}

We finally focus on how the models perform across different entity types: organizations, locations, and persons. This analysis considers only the after scenario, examining how the models handle various types of events beyond their training cut-off date. The results reveal no clear
preference for any specific type. Instead, the results vary between models. Some models performed consistently across organizations, locations, and persons, while others showed more variability. This suggests that model performance is influenced by their internal configurations and training rather than by the type of entity.

%% file: tables/aff_vs_lik_before.tex
\begin{table}[H]
    \centering
    \captionsetup{justification=centering, singlelinecheck=false}   
    \small
    \resizebox{\columnwidth}{!}{%
    \begin{tabular}{lcccc|cccc}
        \toprule
        Model & \multicolumn{4}{c}{Affirmative} & \multicolumn{4}{c}{Likelihood} \\
        \cmidrule(lr){2-5} \cmidrule(lr){6-9}
        & Prec & Rec & F1 & Acc & Prec & Rec & F1 & Acc \\
        \midrule
        Llama2 7b & 0.651 & \textbf{0.749} & 0.693 & 0.633 & \textbf{0.671} & 0.721 & \textbf{0.694} & \textbf{0.647} \\
        Llama2 70b & 0.964 & 0.069 & 0.128 & 0.478 & \textbf{0.968} & \textbf{0.099} & \textbf{0.179} & \textbf{0.494} \\
        Gemma 7b & 0.758 & \textbf{0.614} & \textbf{0.677} & \textbf{0.673} & \textbf{0.796} & 0.517 & 0.623 & 0.655 \\
        GPT 3.5 Turbo & 0.97 & \textbf{0.125} & \textbf{0.219} & 0.509 & \textbf{0.991} & 0.122 & 0.214 & 0.509 \\
        Mistral 7b & 0.947 & \textbf{0.129} & \textbf{0.226} & \textbf{0.509} & \textbf{0.983} & 0.121 & 0.215 & 0.508 \\
        Mixtral 8x7b & 0.538 & 0.024 & 0.046 & 0.437 & \textbf{0.592} & \textbf{0.06} & \textbf{0.108} & \textbf{0.445} \\
        \bottomrule
    \end{tabular}
    }
    \caption{Performance comparison of \textbf{Affirmative vs. Likelihood} metrics for the before cut-off events.}
    \label{tab:bef_aff_vs_lik}
\end{table}

%% file: tables/aff_vs_lik_after.tex
\begin{table}[H]
    \centering
    \captionsetup{justification=centering, singlelinecheck=false}
    \small
    \resizebox{\columnwidth}{!}{%
    \begin{tabular}{lcccc|cccc}
        \toprule
        Model & \multicolumn{4}{c}{Affirmative} & \multicolumn{4}{c}{Likelihood} \\
        \cmidrule(lr){2-5} \cmidrule(lr){6-9}
        & Prec & Rec & F1 & Acc & Prec & Rec & F1 & Acc \\
        \midrule
        Llama2 7b & 0.660 & \textbf{0.718} & 0.682 & 0.617 & \textbf{0.687} & 0.711 & \textbf{0.696} & \textbf{0.642} \\
        Llama2 70b & 0.932 & 0.046 & 0.087 & 0.446 & \textbf{0.954} & \textbf{0.061} & \textbf{0.113} & \textbf{0.454} \\
        Gemma 7b & 0.723 & \textbf{0.470} & \textbf{0.561} & 0.584 & \textbf{0.782} & 0.438 & 0.554 & \textbf{0.600} \\
        GPT 3.5 Turbo & 0.952 & \textbf{0.059} & \textbf{0.109} & \textbf{0.452} &\textbf{ 0.977} & 0.044 & 0.083 & 0.445 \\
        Mistral 7b & 0.862 & 0.033 & 0.063 & 0.436 & \textbf{0.958} & \textbf{0.041} & \textbf{0.077} & \textbf{0.0443}\\
        Mixtral 8x7b & \textbf{0.733} & 0.044 & 0.082 & 0.433 & 0.689 & \textbf{0.084} & \textbf{0.148} & \textbf{0.442} \\
        \bottomrule
    \end{tabular}
    }
    \caption{Performance comparison of \textbf{Affirmative vs. Likelihood} metrics for the after cut-off events.}
    \label{tab:aft_aff_vs_lik}
\end{table}

%% file: tables/reasoning.tex
\begin{table}[H]
    \centering
    \captionsetup{justification=centering, singlelinecheck=false}
    \small
    \resizebox{\columnwidth}{!}{%
    \begin{tabular}{lcccc|cccc}
        \toprule
        Model & \multicolumn{4}{c}{Likelihood} & \multicolumn{4}{c}{Likelihood + Reasoning} \\
        \cmidrule(lr){2-5} \cmidrule(lr){6-9}
        & Prec & Rec & F1 & Acc & Prec & Rec & F1 & Acc \\
        \midrule
        Llama2 7b & 0.727 & 0.723 & 0.720 & 0.654 & \textbf{0.695} & \textbf{0.921} & \textbf{0.790} & \textbf{0.695} \\
        Llama2 70b & \textbf{0.976} & \textbf{0.055} & \textbf{0.102} & \textbf{0.413} & 0.000 & 0.000 & 0.000 & 0.381 \\
        Gemma 7b & \textbf{0.835} & \textbf{0.454} & \textbf{0.578} & \textbf{0.601} & 0.754 & 0.403 & 0.521 & 0.546 \\
        GPT 3.5 Turbo & \textbf{0.979} & 0.044 & 0.082 & 0.406 & 0.932 & \textbf{0.114} & \textbf{0.199} & \textbf{0.444} \\
        Mistral 7b & \textbf{1.000} & 0.030 & 0.057 & 0.399 & 0.885 & \textbf{0.159} & \textbf{0.266} & \textbf{0.467} \\
        Mixtral 8x7b & 0.588 & 0.075 & 0.132 & 0.397 & \textbf{1.000} & \textbf{0.025} & \textbf{0.049} & 0.397 \\
        \bottomrule
    \end{tabular}
    }
    \caption{Performance comparison of \textbf{Likelihood Approach} and \textbf{Reasoning Approach} metrics for the after cut-off events.}
    \label{tab:lik_and_reason}
\end{table}

%% file: tables/counterfactual.tex
\begin{table}[H]
    \centering
    \captionsetup{justification=centering, singlelinecheck=false}
    \small
    \resizebox{\columnwidth}{!}{%
    \begin{tabular}{lcccc|cccc}
        \toprule
        Model & \multicolumn{4}{c}{Positive instances} & \multicolumn{4}{c}{Counterfactual instance} \\
        \cmidrule(lr){2-5} \cmidrule(lr){6-9}
        & Prec & Rec & F1 & Acc & Prec & Rec & F1 & Acc \\
        \midrule
        Llama2 7b & \textbf{0.709} & \textbf{0.700} & \textbf{0.699} & \textbf{0.636} & 0.333 & 0.013 & 0.026 & 0.379 \\
        Gemma 7b & \textbf{0.817} & \textbf{0.478} & \textbf{0.574} & \textbf{0.590} & 0.7884 & 0.185 & 0.288 & 0.448 \\
        GPT 3.5 Turbo & \textbf{0.750} & \textbf{0.041} & \textbf{0.077} & \textbf{0.400} & 0.000 & 0.000 & 0.000 & 0.379 \\
        \bottomrule
    \end{tabular}
    }
    \caption{Performance comparison between \textbf{Positive instances} and \textbf{Counterfactual instances} metrics.}
    \label{tab:pos_vs_counter}
\end{table}

%% file: sections/conclusion.tex
Future forecasting is daily activity of everyone. %Humans try to continuously predict the future.  
However, the forecasting abilities of LLMs have still not been adequately explored. In this research, we explored the predictive capabilities of various language models, focusing on their performance in different scenarios and approaches. The evaluations included different questioning approaches, temporal ranges, popularity, and entity types. The results showed that the Likelihood approach showed a slight edge in the affirmative vs. likelihood scenario, yet similar performances in the before and after cut-off training date scenarios. The Reasoning approach achieved higher recall rates yet with an increased rate of false positives, indicating a tendency to classify more events as positive. The Counterfactual approach highlighted a decline in performance, suggesting sensitivity to slight changes.

Our research contributes to the field of predictive modeling using LLMs, focusing on predictive tasks across various scenarios. We created a time-sensitive dataset for future prediction tasks, which serves as a basis for examining LLMs' predictive capabilities and identifying areas needing further research. The dataset includes a diverse set of questioning scenarios providing a comprehensive view of LLMs' performance across different types of predictive tasks, which are all temporally aligned with the cut-off dates of the tested models. The study also analyzed the dataset in different scenarios, to explore potential correlations between these factors and the models' performance.

Our future research will focus on extending data collection and investigating event plausibility, refining temporal reasoning, and exploring models' sensitivity to slight changes. Investigating ethical considerations and guardrails that might affect predictions, along with automated predictive text generation, would further enhance the understanding and application of LLMs in predictive modeling tasks.